\documentclass[runningheads]{llncs}
\usepackage{graphicx}

\usepackage{tikz}
\usepackage{comment}
\usepackage{amsmath,amssymb} 
\usepackage{color}
\usepackage{multirow}
\newcommand{\etal}{\textit{et al.}}
\newcommand{\ie}{\textit{i.e. }}
\newcommand{\eg}{\textit{e.g.}}

\usepackage{booktabs}
\usepackage{listings}
\usepackage{xcolor}

\definecolor{codegreen}{rgb}{0,0.6,0}
\definecolor{codegray}{rgb}{0.5,0.5,0.5}
\definecolor{codepurple}{rgb}{0.58,0,0.82}
\definecolor{backcolour}{rgb}{0.95,0.95,0.92}

\usepackage[accsupp]{axessibility}  

\begin{document}
\pagestyle{headings}
\mainmatter
\def\ECCVSubNumber{3593}  

\title{Discovering Human-Object Interaction Concepts via Self-Compositional Learning} 

\titlerunning{Discovering Human-Object Interaction Concepts}
%
\author{Zhi Hou\inst{1} \and
Baosheng Yu\inst{1} \and
Dacheng Tao\inst{1,2}}
\authorrunning{Z. Hou, B. Yu et al.}
%
\institute{
The University of Sydney, Australia \\
\and
JD Explore Academy, China\\
\email{\tt\small zhou9878@uni.sydney.edu.au, baosheng.yu@sydney.edu.au,dacheng.tao@gmail.com}
}



\maketitle
\begin{abstract}
A comprehensive understanding of human-object interaction (HOI) requires detecting not only a small portion of predefined HOI concepts (or categories) but also other reasonable HOI concepts, while current approaches usually fail to explore a huge portion of unknown HOI concepts (i.e., unknown but reasonable combinations of verbs and objects). In this paper, 1) we introduce a novel and challenging task for a comprehensive HOI understanding, which is termed as \textbf{HOI Concept Discovery}; and 2) we devise a self-compositional learning framework (or \textbf{SCL}) for HOI concept discovery. Specifically, we maintain an online updated concept confidence matrix during training: 1) we assign pseudo labels for all composite HOI instances according to the concept confidence matrix for self-training; and 2) we update the concept confidence matrix using the predictions of all composite HOI instances. Therefore, the proposed method enables the learning on both known and unknown HOI concepts. We perform extensive experiments on several popular HOI datasets to demonstrate the effectiveness of the proposed method for HOI concept discovery, object affordance recognition and HOI detection. For example, the proposed self-compositional learning framework significantly improves the performance of 1) HOI concept discovery by over \textbf{10\%} on HICO-DET and over \textbf{3\%} on V-COCO, respectively; 2) object affordance recognition by over \textbf{9\%} mAP on MS-COCO and HICO-DET; and 3) rare-first and non-rare-first unknown HOI detection relatively over \textbf{30\%} and \textbf{20\%}, respectively. Code is publicly available at \url{https://github.com/zhihou7/HOI-CL}.


\end{abstract}

\section{Introduction}
\label{sec:intro}

Human-object interaction (HOI) plays a key role in analyzing the relationships between humans and their surrounding objects~\cite{gupta2009observing}, which is of great importance for deep understanding on human activities/behaviors. Human-object interaction understanding has attracted extensive interests from the community, including image-based~\cite{chaoy2015hico,chaoy2018learning,gao2018ican,liao2019ppdm,tamura_cvpr2021}, video-based visual relationship analysis~\cite{damen2018scaling1,materzynska2020something}, video generation~\cite{nawhal2020generating}, and  scene reconstruction~\cite{zhang2020phosa}. However, the distribution of HOI samples is naturally long-tailed: most interactions are rare and some interactions do not even occur in most scenarios, since we can not obtain an interaction between human and object until someone conducts such action in real-world scenarios. Therefore, recent HOI approaches mainly focus on the analysis of very limited predefined HOI concepts/categories, leaving the learning on a huge number of unknown HOI concepts~\cite{coren2003sensation,best1986cognitive}  poorly investigated, including HOI detection and object affordance recognition~\cite{shen2018scaling,hou2020visual,hou2021atl}. For example, there are only 600 HOI categories known in HICO-DET~\cite{chao2018learning}, while we can find 9,360 possible verb-object combinations from $117$ verbs and $80$ objects.

\begin{figure}
  \centering
  \includegraphics[width=0.73\linewidth]{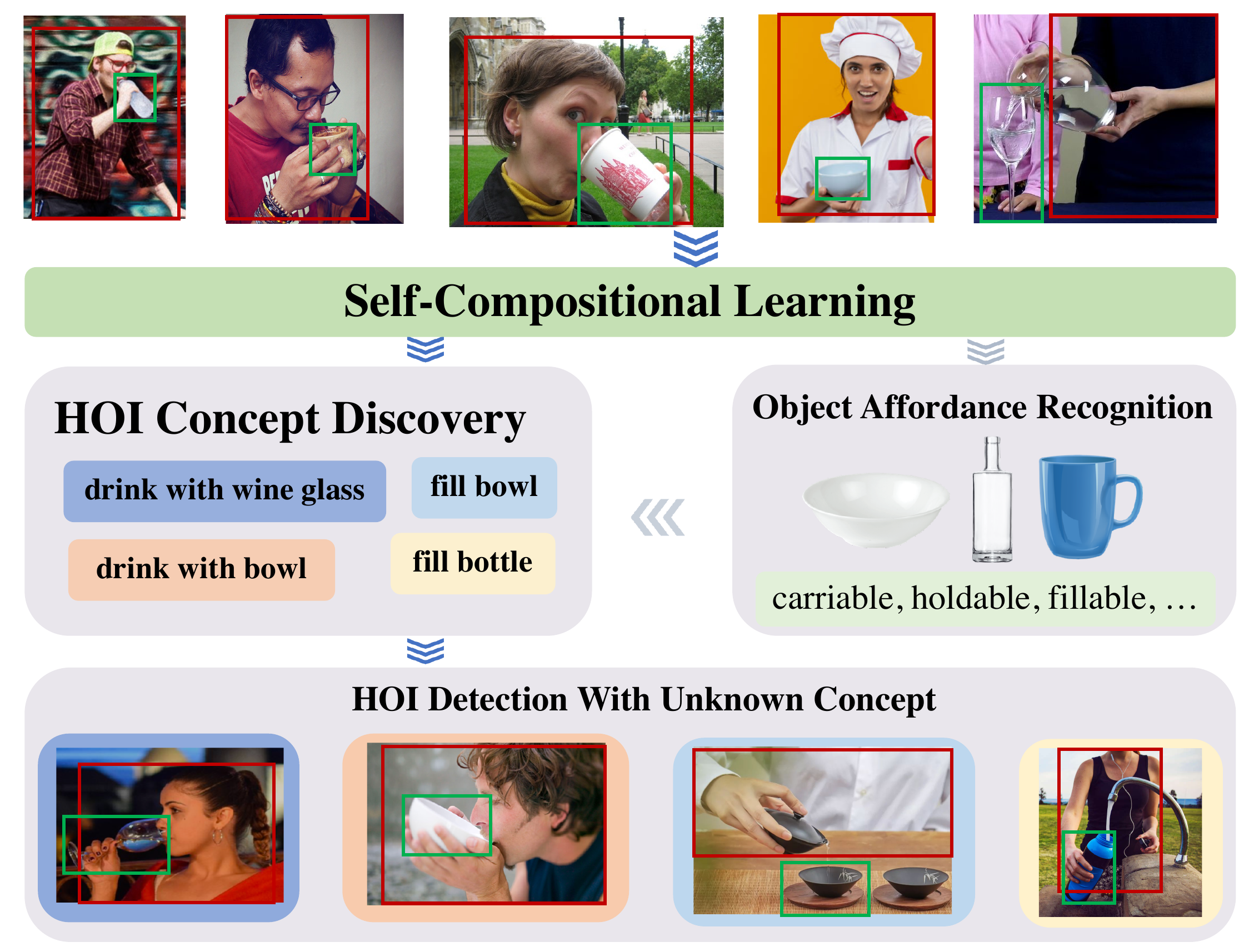}
  \caption{An illustration of unknown HOI detection via  concept discovery. Given some known HOI concepts (\eg, ``drink$\_$with cup", ``drink$\_$with bottle", and ``hold bowl"), the task of concept discovery aims to identify novel HOI concepts (i.e., reasonable combinations between verbs and objects). For example, here we have some novel HOI concepts, ``drink$\_$with wine$\_$glass", ``fill bowl", and ``fill bottle". Specifically, the proposed self-compositional learning framework jointly optimizes HOI concept discovery and HOI detection on unknown concepts in an end-to-end manner.
  }
  \label{fig:concpet}
\end{figure}

Object affordance is closely related to HOI understanding from an object-centric perspective. Specifically, two objects with similar attributes usually share the same affordance, \ie, humans usually interact with similar objects in a similar way~\cite{gibson1979The}. For example, cup, bowl, and bottle share the same attributes (\eg, hollow), and all of these objects can be used to ``drink with".
Therefore, object affordance~\cite{gibson1979The,hou2021atl} indicates whether each action can be applied into an object, \ie, if a verb-object combination is reasonable, we then find a novel HOI concept/category. An illustration of unknown HOI detection via concept discovery is shown in Fig.~\ref{fig:concpet}. Recently, it has turned out that an HOI model is not only capable of detecting interactions, but also able to recognize object affordances~\cite{hou2021atl}, especially novel object affordances using the composite HOI features. Particularly, novel object affordance recognition also indicates discovering novel reasonable verb-object combinations or HOI concepts. Inspired by this, we  can introduce a simple baseline for HOI concept discovery by averaging the affordance predictions of training dataset into each object category~\cite{hou2021atl}.

Nevertheless, there are two main limitations when directly utilizing object affordance prediction~\cite{hou2021atl} for concept discovery. First, the affordance prediction approach in \cite{hou2021atl} is time-consuming and unsuitable to be utilized during training phrase, since it requires to predict all possible combinations of verbs and objects using the whole training set. By contrast, we introduce an online HOI concept discovery method, which is able to collect concept confidence in a running mean manner with verb scores of all composite features in mini-batches during training. Second, also more importantly, the compositional learning approach~\cite{hou2021atl} merely optimizes the composite samples with known concepts (\eg, 600 categories on HICO-DET), ignoring a large number of  composite samples with unknown concepts (unlabeled composite samples). As a result, the model is inevitably biased to known object affordances (or HOI concepts), and leads to the similar inferior performance to the one in Positive-Unlabeled learning~\cite{de1999positive,elkan2008learning,scott2009novelty}. That is, without negative samples for training, the network will tend to predict high confidence on those impossible verb-object combinations or overfit verb patterns (please refer to Appendix A for more analysis). Considering that the online concept discovery branch is able to predict concept confidence during optimization, we can then construct pseudo labels~\cite{lee2013pseudo} for all composite HOIs belonging to either known or unknown categories. Inspired by this, we introduce a self-compositional learning strategy (or SCL) to jointly optimize all composite representations and improve concept predictions in an iterative manner. Specifically, SCL combines the object representations with different verb representations to compose new samples for optimization, and thus implicitly pays attention to the object representations and improves the discrimination of composite representations. By doing this, we can improve the object affordance learning, and then facilitate the HOI concept discovery. 

Our main contributions can be summarized as follows: 1) we introduce a new task for a better and comprehensive understanding on human-object interactions; 2) we devise a self-compositional learning framework for HOI concept discovery and object affordance recognition simultaneously; and 3) we evaluate the proposed approach on two extended benchmarks, and it significantly improves the performance of HOI concept discovery, facilitates object affordance recognition with HOI model, and also enables HOI detection with novel concepts.

\section{Related Work}

\subsection{Human-Object Interaction}
HOI understanding~\cite{gupta2009observing} is of great importance for visual relationship reasoning~\cite{xu2017scene} and action understanding~\cite{carreira2017quo,zheng2020skeleton}. Different approaches have been investigated for HOI understanding from various aspects, including HOI detection~\cite{chao2018learning,li2018transferable,liao2019ppdm,zhong2020polysemy,kim2021hotr,chen2021reformulating,zou2021end,tamura_cvpr2021,zhang2021mining}, HOI recognition~\cite{chaoy2015hico,kato2018compositional,huynh2021interaction}, video HOI~\cite{damen2018scaling1,ji2021detecting}, compositional action  recognition~\cite{materzynska2020something}, 3D scene reconstruction~\cite{zhang2020phosa,dabral2021gravity}, video generation~\cite{nawhal2020generating}, and object affordance reasoning~\cite{fang2018demo2vec,hou2021atl}. Recently, compositional approaches (\eg, VCL~\cite{hou2020visual}) have been intensively proposed for HOI understanding using the structural characteristic~\cite{kato2018compositional,hou2020visual,nawhal2020generating,li2020hoi,hou2021atl}. Meanwhile, DETR-based methods (\eg, Qpic~\cite{tamura_cvpr2021}) achieve superior performance on HOI detection. However, these approaches mainly consider the perception of known HOI concepts, and pay no attention to HOI concept discovery. To fulfill the gap between learning on known and unknown concepts, a novel task, \ie, HOI concept discovery, is explored in this paper. Currently, zero-shot HOI detection also attracts massive interests from the community~\cite{shen2018scaling,bansal2020detecting,preye2019detecting,hou2020visual,hou2021fcl}. However, those approaches merely consider known concepts and are unable to discover HOI concepts. Some HOI approaches~\cite{preye2019detecting,bansal2020detecting,wang2020discovering,wang2021discovering} expand the known concepts via leveraging language priors. However, that is limited to existing knowledge and can not discover concepts that never appear in the language prior knowledge. HOI concept discovery is able to address the problem, and enable unknown HOI concept detection.






\subsection{Object Affordance Learning}

The notation of affordance is formally introduced in ~\cite{gibson1979The}, where object affordances are usually those action possibilities that are perceivable by an actor~\cite{norman2002the,gibson1979The,gibson2014ecological}. Noticeably, the action possibilities of an object also indicate the HOI concepts related to the object. Therefore, object affordance can also represent the existence of HOI concepts. Recent object affordance approaches mainly focus on the pixel-level affordance learning from human interaction demonstration~\cite{kjellstrom2011visual,Fouhey2014PeopleWH,fang2018demo2vec,hassan2015attribute,nagarajan2020learning,deng20213d,zhai2021one}. Yao \etal~\cite{Yao2013Discovering} present a weakly supervised approach to discover object functionalities from HOI data in the musical instrument environment. Zhu \etal~\cite{zhu2014reasoning} introduce to reason affordances in knowledge-based representation. Recent approaches propose to generalize HOI detection to unseen HOIs via functionality generalization~\cite{bansal2020detecting} or analogies~\cite{preye2019detecting}. However those approaches focus on HOI detection, ignoring object affordance recognition. Specifically, Hou \etal~\cite{hou2021atl} introduce an affordance transfer learning (ATL) framework to enable HOI model to not only detect interactions but also recognize object affordances.
Inspired by this, we further develop a self-compositional learning framework to facilitate the object affordance recognition with HOI model to discover novel HOI concepts for downstream HOI tasks.

\subsection{Semi-Supervised Learning}

Semi-supervised learning is a learning paradigm for constructing models that use both labeled and unlabeled data \cite{yang2021survey}. There are a wide variety of Deep Semi-Supervised Learning methods, such as Generative Networks \cite{kingma2014semi,springenberg2015unsupervised}, Graph-Based methods \cite{wang2016structural,gilmer2017neural}, Pseudo-Labeling methods \cite{lee2013pseudo,xie2020self,hinton2015distilling}. HOI concept discovery shares a similar characteristic to semi-supervised learning approaches. {\textit HOI concept discovery has instances of labeled HOI concepts, but no instances of unknown concepts.} We thus compose HOI representations for unknown concepts according to~\cite{scudder1965probability}. With composite HOIs, concept discovery and object affordance recognition can be treated as PU learning~\cite{de1999positive}. Moreover, HOI concept discovery requires to discriminate whether the combinations (possible HOI concepts) are reasonable and existing. Considering each value of the concept confidences also represents the possibility of the composite HOI, we construct pseudo labels~\cite{lee2013pseudo,scudder1965probability} for composite features from the concept confidence matrix, and optimize the composite HOIs in an end-to-end way.


\section{Approach}

In this section, we first formulate the problem of HOI concept discovery and introduce the compositional learning framework. We then describe a baseline for HOI concept discovery via affordance prediction. Lastly, we introduce the proposed self-compositional learning framework for online HOI concept discovery and object affordance recognition.

\begin{figure*}[!ht]
  \centering
  \includegraphics[width=0.9\linewidth]{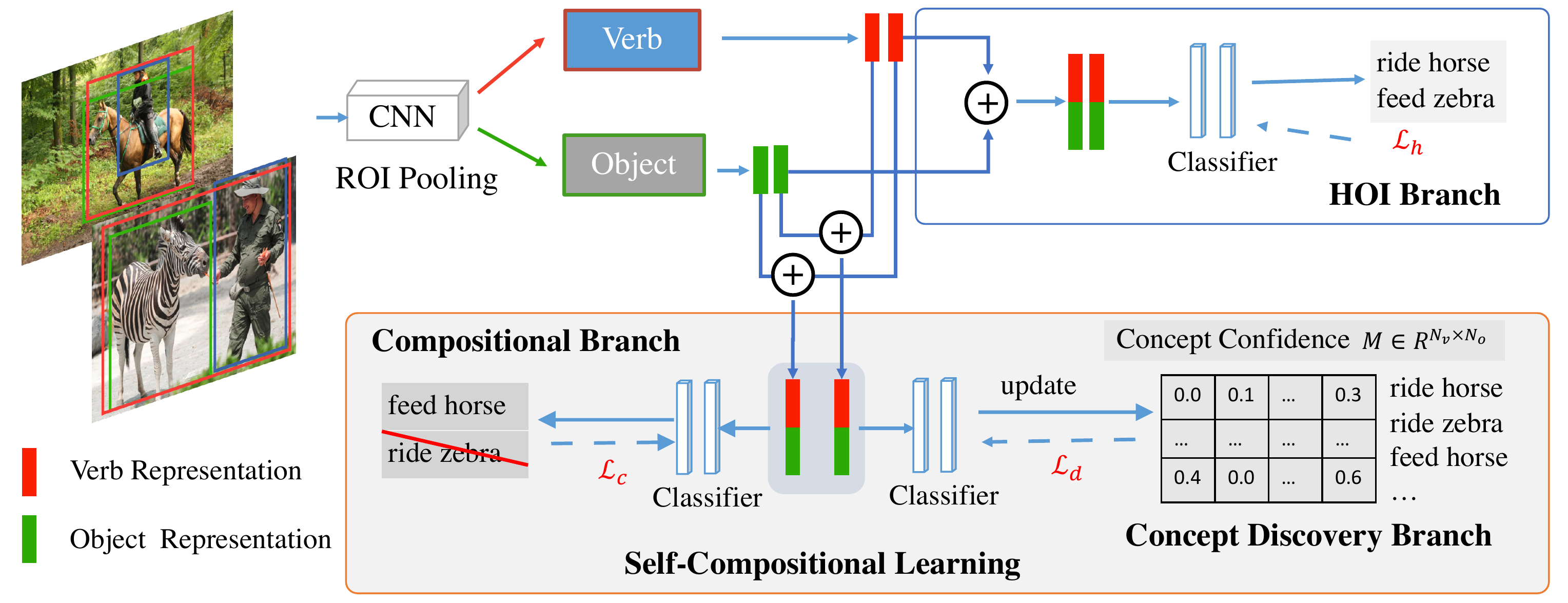}
  \caption{Illustration of Self-Compositional Learning for HOI Concept Discovery. Specifically, following~\cite{hou2020visual}, verb and object features are extracted via RoI-Pooling from union box and object box respectively, which are then used to construct HOI features in HOI branch according to HOI annotation. Following~\cite{hou2020visual}, for SCL, verb and object features are further mutually combined to generate composite HOI features. Then, the feasible composite HOI features belonging to the known concepts are directly used to train the network in Compositional Branch. Here the classifier predicts verb classes directly. Meanwhile, we update the concept confidence $\mathbf{M} \in R^{N_v \times N_o}$, where $N_v$ and $N_o$ are the number of verb classes and object classes respectively, with the predictions of all composite HOI features. The concept discovery branch is optimized via a self-training approach to learn from composite HOI features with the concept confidence $\mathbf{M}$.}
  \label{fig:overview}
\end{figure*}


\subsection{Problem Definition}

HOI concept discovery aims to discover novel HOI concepts/categories using HOI instances from existing known HOI categories. Given a set of verb categories $\mathcal{V}$ and a set of object categories $\mathcal{O}$, let $\mathcal{S}=\mathcal{V}\times \mathcal{O}$ indicate the set of all possible verb-object combinations. Let $\mathcal{S}^{k}$, $\mathcal{S}^{u}$, and $\mathcal{S}^{o}$ denote three disjoint sets, known HOI concepts, unknown HOI concepts, and invalid concepts (or impossible verb-object combinations), respectively. That is, we have $\mathcal{S}^{k} \cap \mathcal{S}^{u}=\varnothing$ and $\mathcal{S}^k \cup \mathcal{S}^u = \mathcal{S}$ if $\mathcal{S}^o=\varnothing$. Let $\mathcal{T}=\{(h_i, c_i)\}_{i=1}^{L}$ indicate the training dataset, where $h_i$ is a HOI instance (\ie, verb-object visual representation pair), $c_i \in \mathcal{S}^k$ indicates the label of the $i$-th HOI instance and $L$ is the total number of HOI instance.

We would also like to clarify the difference between the notations of ``unknown HOI categories" and ``unseen HOI categories" in current HOI approaches as follows.
Let $\mathcal{S}^{z}$ indicate the set of ``unseen HOI categories" and we then have $\mathcal{S}^{z} \subseteq \mathcal{S}^{k}$.
Specifically, ``unseen HOI category" indicates that the HOI concept is known but no corresponding HOI instances can be observed in the training data. Current HOI methods usually assume that unseen HOI categories $\mathcal{S}^{z}$ are known HOI categories via the prior knowledge~\cite{shen2018scaling,kato2018compositional,preye2019detecting,bansal2020detecting,hou2020visual}. Therefore, existing HOI methods can not directly detect/recognize HOIs with unknown HOI concepts. HOI concept discovery aims to find $\mathcal{S}^u$ from the existing HOI instances in $\mathcal{T}$ with only known HOI concepts in $\mathcal{S}^{k}$.

\subsection{HOI Compositional Learning}

Inspired by the compositional nature of HOI, \ie, each HOI consists of a verb and an object, visual compositional learning has been intensively explored for HOI detection by combining visual verb and object representations~\cite{kato2018compositional,hou2020visual,hou2021fcl,hou2021atl}. Let $\mathbf{h}_i = \langle \mathbf{x}_{v_i}, \mathbf{x}_{o_i} \rangle$ indicate a HOI instance, where $\mathbf{x}_{v_i}$ and $\mathbf{x}_{o_i}$ denote the verb and object representations, respectively. The HOI compositional learning then aims to achieve the following objective,
\begin{equation}
   g_{h}(\langle\widetilde{\mathbf{x}}_{v_i},\widetilde{\mathbf{x}}_{o_i}\rangle) \approx  g_{h}(\langle\mathbf{x}_{v_i},\mathbf{x}_{o_i}\rangle),
\end{equation}
where $g_{h}$ indicates the HOI classifier, $\mathbf{x}_{v_i}$ and $\mathbf{x}_{o_i}$ indicate the real verb-object representation pair (\ie, annotated HOI pair in dataset),  $\langle \widetilde{\mathbf{x}}_{v_i}, \widetilde{\mathbf{x}}_{o_i}\rangle$ indicates the composite verb-object pair. Specifically, $\widetilde{\mathbf{x}}_{o_i}$ can be obtained from either real HOIs~\cite{hou2020visual}, fabricated objects or language embedding~\cite{hou2021fcl,bansal2020detecting,preye2019detecting}, or external object datasets~\cite{hou2021atl}, while $\widetilde{\mathbf{x}}_{v_i}$ can be from real HOIs (annotated verb-object pair) and language embeddings~\cite{kato2018compositional,preye2019detecting}. As a result, when composite HOIs are similar to real HOIs, we are then able to augment HOI training samples in a compositional manner. However, current compositional approaches for HOI detection~\cite{hou2020visual,hou2021atl} simply remove the composite HOI instances out of the label space, which may also remove a large number of feasible HOIs (\eg, ``ride zebra" as shown Figure~\ref{fig:overview}). Furthermore, the compositional approach can not only augment the training data for HOI recognition, but also provide a method to determinate whether $\widetilde{\mathbf{x}}_{v_i}$ and $\widetilde{\mathbf{x}}_{o_i}$ are combinable to form a new HOI or not \cite{hou2021atl}, \ie, discovering the HOI concepts.

\subsection{Self-Compositional Learning}

In this subsection, we introduce the proposed self-compositional learning framework for HOI concept discovery as follows. As shown in Figure~\ref{fig:overview}, the main HOI concept discovery framework falls into the popular two-stage HOI detection framework~\cite{hou2020visual}.
Specifically, we compose novel HOI samples from pair-wise images to optimize the typical HOI branch (annotated HOIs), compositional branch (the composite HOIs out of the label space are removed~\cite{hou2020visual,hou2021atl}) and the new concept discovery branch (all composite HOIs are used).
The main challenge of HOI concept discovery is the lack of instances for unknown HOI concepts, but we can infer to discover new concepts according to the shared verbs and objects. Specifically, we find that the affordance transfer learning~\cite{hou2021atl} can be used for not only the object affordance recognition but also the HOI concept discovery, and we thus first introduce the affordance-based method as a baseline as follows.

\subsubsection{Affordance Prediction}
\label{sec:aff_pred}
The affordance transfer learning~\cite{hou2021atl} or ATL is introduced for affordance recognition using the HOI detection model. However, it has been ignored that the affordance prediction can also enable HOI concept discovery, \ie, predicting a new affordance for an object although the affordance is not labeled during training. We describe a vanilla approach for HOI concept discovery using affordance prediction~\cite{hou2021atl}. Specifically, we predict the affordances for all objects in the training set according to ~\cite{hou2021atl}. Then, we average the affordance predictions according to each object category to obtain the HOI concept confidence matrix $\mathbf{M} \in R^{N_v \times N_o}$, where each value represents the concept confidence of the corresponding combination between a verb and an object. $N_v$ and $N_o$ are the numbers of verb and object categories, respectively. For simplicity, we may use both vector and matrix forms of the confidence matrix $\mathbf{M} \in R^{N_vN_o}$ and $\mathbf{M} \in R^{N_v \times N_o}$ in this paper. Though affordance prediction can be used for HOI concept discovery, it is time-consuming since it requires to predict affordances of all objects in training set. Specifically, we need an extra offline affordance prediction process to infer concepts with the computational complexity $O(N^2)$ in~\cite{hou2021atl}, where $N$ is the number of total training HOIs, \eg, it takes 8 hours with one GPU to infer the concept matrix $\mathbf{M}$ on HICO-DET. However, we can treat the verb representation as affordance representation~\cite{hou2021atl}, and obtain the affordance predictions for all objects in 
each mini-batch during training stage. Inspired by the running mean manner in~\cite{ioffe2015batch}, we devise an online HOI concept discovery framework via averaging the predictions in each mini-batch.

\subsubsection{Online Concept Discovery}
 As shown in Figure~\ref{fig:overview}, we keep a HOI concept confidence vector during training, $\mathbf{M} \in R^{N_vN_o}$, where each value represents the concept confidence of the corresponding combination between a verb and an object. To achieve this, we first extract all verb and object representations among pair-wise images in each batch as $\mathbf{x}_v$ and $\mathbf{x}_o$. We then combine each verb representation and all object representations to generate the composite HOI representations $\mathbf{x}_{h}$. After that, we use the composite HOI representations as the input to the verb classifier and obtain the corresponding verb predictions $\mathbf{\hat{Y}}_v\in R^{NN\times N_v}$, where $N$ indicates the number of real HOI instances (\ie, verb-object pair) in each mini-batch and $NN$ is then the number of all composite verb-object pairs (including unknown HOI concepts). Let $\mathbf{Y}_v \in R^{N \times N_v}$ and $\mathbf{Y}_o \in R^{N \times N_o}$ denote the label of verb representations $\mathbf{x}_v$ and object representations $\mathbf{x}_o$, respectively. We then have all composite HOI labels $\mathbf{Y}_{h} = \mathbf{Y}_v \otimes \mathbf{Y}_o$, where $\mathbf{Y}_{h} \in R^{ N N\times N_v N_o}$, and the superscripts $h$, $v$, and $o$ indicate HOI, verb, and object, respectively. Similar to affordance prediction, we repeat $\mathbf{\hat{Y}}_v$ by $N_o$ times to obtain concept predictions $\mathbf{\hat{Y}}_{h} \in R^{N N \times N_v N_o}$. Finally, we update $\mathbf{M}$ in a running mean manner \cite{ioffe2015batch} as follows,
\begin{align}
    \mathbf{M} &\leftarrow \frac{\mathbf{M} \odot \mathbf{C} + \sum_{i}^{NN}{\mathbf{\hat{Y}}_{h}(i,:) \odot \mathbf{Y}_{h}(i,:)}}{\mathbf{C}+\sum_{i}^{NN}{\mathbf{Y}_{h}(i,:)}}, \\
    \mathbf{C} &\leftarrow \mathbf{C} + \sum_{i}^{NN}{\mathbf{Y}_{h}(i,:)},
\end{align}
where $\odot$ indicates the element-wise multiplication, $\mathbf{\hat{Y}}_{h}(i,:) \odot \mathbf{Y}_{h}(i,:)$ aims to filter out predictions whose labels are not $\mathbf{Y}_{h}(i,:)$, each value of $\mathbf{C} \in R^{N_v N_o}$ indicates the total number of composite HOI instances in each verb-object pair (including unknown HOI categories). Actually, $\mathbf{\hat{Y}}_{h}(i,:) \odot \mathbf{Y}_{h}(i,:)$ follows the affordance prediction process~\cite{hou2021atl}. The normalization with $\mathbf{C}$ is to avoid the model bias to frequent categories. Specifically, both $\mathbf{M}$ and $\mathbf{C}$ are zero-initialized. With the optimization of HOI detection, we can obtain the vector $\mathbf{M}$ to indicate the HOI concept confidence of each combination between verbs and objects.

\subsubsection{Self-Training}

Existing HOI compositional learning approaches~\cite{hou2020visual,hou2021fcl,hou2021atl} usually only consider the known HOI concepts and simply discard the composite HOIs out of label space during optimization. Therefore, there are only positive data for object affordance learning, leaving a large number of unlabeled composite HOIs ignored. Considering that the concept confidence on HOI concept discovery also demonstrates the confidence of affordances (verbs) that can be applied to an object category, we thus try to explore the potential of all composite HOIs, i.e., both labeled and unlabeled composite HOIs,  in a semi-supervised way. Inspired by the way used in PU learning~\cite{de1999positive} and pseudo-label learning~\cite{lee2013pseudo}, we devise a self-training strategy by assigning the pseudo labels to each verb-object combination instance using the concept confidence matrix $\mathbf{M}$, and optimize the network with the pseudo labels in an end-to-end way. With the self-training, the online concept discovery can gradually improve the concept confidence $\mathbf{M}$, and in turn optimize the HOI model for object affordance learning with the concept confidence.
Specifically, we construct the pseudo labels $\mathbf{\tilde{Y}}_v \in R ^{NN \times N_v}$ from the concept confidence matrix $\mathbf{M} \in R^{N_v \times N_o}$ for composite HOIs $\mathbf{x}_{h}$ as follows,


\begin{equation}
\label{eq:pseudo}
    \mathbf{\tilde{Y}}_v(i,:) = \sum_j^{N_o} \frac{\mathbf{M}(:,j)}{\max(\mathbf{M})} \odot \mathbf{Y}_h(i,:,j),
\end{equation}
where $0 \leq j < N_o$ indicates the index of object category, $0 \leq i < NN$ is the index of HOI representations. Here, $N$ is the number of HOIs in each mini-batch, and is usually very small on HICO-DET and V-COCO. Thus the time complexity of Equation~\ref{eq:pseudo} is small. The labels of composite HOIs are reshaped as $\mathbf{Y}_h \in R^{NN \times N_v \times N_o}$. Noticeably, in each label $\mathbf{Y}_{h}(i, :, :)$, there is only one vector $\mathbf{Y}_{h}(i,:,j)$ larger than 0 because each HOI has only one object. As a result, we obtain pseudo verb label $\mathbf{\tilde{Y}}_{v}(i,:)$ for HOI $\mathbf{x}_{h_i}$. Finally, we use composite HOIs with pseudo labels to train the models, and the loss function is defined as follows,
\begin{equation}
    \mathcal{L}_d = \frac{1}{NN} \sum_i^{NN} (\frac{1}{N_v} \sum_{k}^{N_v} \mathcal{L}_{\text{BCE}} (\frac{\mathbf{Z}(i,k)}{T}, \mathbf{\tilde{Y}}_{v}(i,k))),
\end{equation}
where $\mathbf{Z}(i,:)$ is the prediction of the $i$-th composite HOI, $0 \leq k<N_v$ means the index of predictions, $T$ is the temperature hyper-parameter to smooth the predictions (the default value is 1 in experiment), $\mathcal{L}_{\text{BCE}}$ indicates the binary cross entropy loss. Finally, we optimize the network using
$\mathcal{L}_d$, $\mathcal{L}_{h}$ and $\mathcal{L}_{c}$ in an end-to-end way, where $\mathcal{L}_{h}$ indicate the typical classification loss for known HOIs and $\mathcal{L}_{c}$ is the compositional learning loss~\cite{hou2020visual}.

\section{Experiments}
\label{sec:exp}

In this section, we first introduce the datasets and evaluation metrics. We then compare the baseline and the proposed method for HOI concept discovery and object affordance recognition. We also demonstrate the effectiveness of the proposed method for HOI detection with unknown concepts and zero-shot HOI detection. Lastly, we provide some visualizations results of self-compositional learning. Moreover, ablation studies and the full results of HOI detection with self-compositional learning are provided in Appendix D, F, respectively.

\subsection{Datasets and Evaluation Metrics}

\textbf{Datasets.} We extend two popular HOI detection datasets, HICO-DET \cite{chao2018learning} and V-COCO \cite{gupta2015visual}, to evaluate the performance of different methods for HOI concept discovery. Specifically, we first manually annotate all the possible verb-object combinations on HICO-DET (117 verbs and 80 objects) and V-COCO (24 verbs and 80 objects). As a result, we obtain 1,681 concepts on HICO-DET and 401 concepts on V-COCO, \ie, 1,681 of 9,360 verb-object combinations on HICO-DET and 401 of 1,920 verb-object combinations on V-COCO are reasonable. Besides, 600 of 1,681 HOI concepts on HICO-DET and 222 of 401 HOI concepts on V-COCO are known according to existing annotations. Thus, the HOI concept discovery task requires to discover the other 1,081 concepts on HICO-DET and 179 concepts on V-COCO. See more details about the annotation process, the statistics of annotations, and the novel HOI concepts in Appendix B.

\textbf{Evaluation Metrics.} HOI concept discovery aims to discover all reasonable combinations between verbs and objects according to existing HOI training samples. We report the performance by using the average precision (AP) for concept discovery and mean AP (or mAP) for object affordance recognition. For HOI detection, we also report the performance using mAP. We follow~\cite{hou2021atl} to evaluate object affordance recognition with HOI model on COCO validation 2017~\cite{lin2014microsoft}, Object 365 validation~\cite{Shao2020Objects365}, HICO-DET test set~\cite{chao2018learning} and Novel Objects from Object 365~\cite{Shao2020Objects365}.

\subsection{Implementation Details}
\label{sec:impl_detail}
We implement the proposed method with TensorFlow~\cite{abadi2016tensorflow}.
During training, we have two HOI images (randomly selected) in each mini-batch and we follow ~\cite{gao2018ican} to augment ground truth boxes via random crop and random shift. We use a modified HOI compositional learning framework, \ie, we directly predict the verb classes and optimize the composite HOIs using SCL. Following ~\cite{hou2020visual,hou2021fcl}, the overall loss function is defined as $\mathcal{L} = \lambda_1 \mathcal{L}_{h} + \lambda_2 \mathcal{L}_{c} + \lambda_3 \mathcal{L}_d$, where $\lambda_1 = 2$, $\lambda_2 = 0.5$, $\lambda_3=0.5$ on HICO-DET, and $\lambda_1 = 0.5$, $\lambda_2 = 0.5$, $\lambda_3=0.5$ on V-COCO, respectively. Following~\cite{hou2021fcl}, we also include a sigmoid loss for verb representation and the loss weight is $0.3$ on HICO-DET.
For self-training, we remove the composite HOIs when its corresponding concept confidence is 0, \ie, the concept confidence has not been updated. If not stated, the backbone is ResNet-101. The Classifier is a two-layer MLP. We train the model for 3.0M iterations on HICO-DET and 300K iterations on HOI-COCO with an initial learning rate of 0.01. For zero-shot HOI detection, we keep human and objects with the score larger than 0.3 and 0.1 on HICO-DET, respectively. See more ablation studies (\eg, hyper-parameters, modules) in Appendix D. Experiments are conducted using a single Tesla V100 GPU (16GB), except for experiments on Qpic~\cite{tamura_cvpr2021}, which uses four V100 GPUs with PyTorch~\cite{NEURIPS2019_9015}.

\subsection{HOI Concept Discovery}


{\bf Baseline and Methods}. We perform experiments to evaluate the effectiveness of our proposed method for HOI concept discovery. For a fair comparison, we build several baselines and methods as follows,
\begin{itemize}
    \item {\bf Random}: we randomly generate the concept confidence to evaluate the performance.
    \item {\bf Affordance}: discover concepts via affordance prediction~\cite{hou2021atl} as described in Sec~\ref{sec:aff_pred}.
    \item {\bf GAT}~\cite{velivckovic2017graph}: build a graph attention network to mine the relationship among verbs during HOI detection, and discover concepts via affordance prediction.
    \item {\bf Qpic}*~\cite{tamura_cvpr2021}: convert verb and object predictions of ~\cite{tamura_cvpr2021} to concept confidence similar as online discovery.
    \item {\bf Qpic*~\cite{tamura_cvpr2021} +SCL}: utilize concept confidence to update verb labels, and optimize the network (Self-Training). Here, we have no composite HOIs.
\end{itemize}
Please refer to the Appendix for more details, comparisons (\eg, re-training, language embedding), and qualitative discovered concepts with analysis.

\begin{table}[!tp]
\small
\caption{The performance of the proposed method for HOI concept discovery. We report all performance using the average precision (AP) (\%). SCL means self-compositional learning. SCL$-$ means online concept discovery without self-training.
}
\label{table:discover}
\centering
\small
\begin{tabular}{@{}lcccc@{}}
\hline
 \multirow{2}{*}{Method} &
\multicolumn{2}{c}{HICO-DET}&\multicolumn{2}{c}{V-COCO}\cr\cline{2-5}
& Unknown (\%) & Known (\%) &  Unknown (\%) & Known (\%) \cr
\hline
Random & 12.52 & 6.56 & 12.53 & 13.54\\


Affordance~\cite{hou2021atl} & 24.38 & 57.92 & 20.91 & 95.71 \\
GAT~\cite{velivckovic2017graph} & 26.35 & 76.05 & 18.35 & 98.09\\
Qpic*~\cite{tamura_cvpr2021} & 27.53 & 87.68 & 15.03 & 13.21\\
\hline

SCL$-$ & 22.25 & 83.04 & 24.89 & 96.70\\
Qpic*~\cite{tamura_cvpr2021} + SCL & 28.44 & 88.91 & 15.48 & 13.34 \\
SCL & {\bf 33.58} & {\bf 92.65} & {\bf 28.77} & {\bf 98.95} \\

\hline
\end{tabular}
\end{table}


{\bf Results Comparison}. Table~\ref{table:discover} shows affordance prediction is capable of HOI concept discovery since affordance transfer learning~\cite{hou2021atl} also transfers affordances to novel objects. Affordance prediction achieves 24.38\% mAP on HICO-DET and 21.36\% mAP on V-COCO, respectively, significantly better than the random baseline. With graph attention network, the performance is further improved a bit. Noticeably, \cite{hou2021atl} completely ignores the possibility of HOI concept discovery via affordance prediction. Due to the strong ability of verb and object prediction, Qpic achieves 27.42\% on HICO-DET, better than affordance prediction. However, Qpic has poor performance on V-COCO. The inference process of affordance prediction for concept discovery is time-consuming (over 8 hours with one GPU). Thus we devise an efficient online concept discovery method which directly predicts all concept confidences. Specifically, the online concept discovery method (SCL$-$) achieves 22.25\% mAP on HICO-DET, which is slightly worse than the result of affordance prediction. On V-COCO, the online concept discovery method improves the performance of concept discovery by {\bf 3.98\%} compared to the affordance prediction. The main reason for the above observation might be due to that V-COCO is a  small dataset and the HOI model can easily overfit known concepts on V-COCO.
Particularly, SCL significantly improves the performance of HOI concept discovery from 22.36\% to {\bf 33.58\%} on HICO-DET and from 24.89\% to {\bf 28.77\%} on V-COCO, respectively. We find we can also utilize self-training to improve concept discovery on Qpic~\cite{tamura_cvpr2021} (ResNet-50) though the improvement is limited, which might be because verbs and objects are entangled with Qpic. Lastly, we meanwhile find SCL largely improves concept discovery of known concepts on both HICO-DET and V-COCO.




\begin{table}[tp]
\caption{Comparison of object affordance recognition with HOI network (trained on HICO-DET) among different datasets. Val2017 is the validation 2017 of COCO \cite{lin2014microsoft}. Obj365 is the validation of Object365 \cite{Shao2020Objects365} with only COCO labels. Novel classes are selected from Object365 with non-COCO labels. ATL$\ast$ means ATL optimized with COCO data. Numbers are copied from the appendix in ~\cite{hou2021atl}. Unknown affordances indicate we evaluate with our annotated affordances. Previous approaches~\cite{hou2020visual,hou2021atl} are usually trained by less 0.8M iterations (Please refer to the released checkpoint in~\cite{hou2020visual,hou2021atl}). We thus also illustrate SCL under 0.8M iterations by default. SCL$-$ means SCL without self-training. Results are reported by Mean Average Precision (\%).}
\label{table:func_obj_ap}
\centering
\small
\begin{tabular}{@{}lc|c|c|c|c|c|c|c@{}}
\hline

\multirow{2}{*}{Method} &
\multicolumn{4}{c}{Known Affordances}&\multicolumn{4}{c}{Unknown Affordances}\cr\cline{2-9}
& Val2017 & Obj365 &HICO & Novel & Val2017 & Obj365 &HICO & Novel \cr

\hline


FCL \cite{hou2021fcl}  &       25.11    &     25.21 &         37.32 &    6.80 & - & - & - & - \\
VCL \cite{hou2020visual}   &     36.74 &  35.73   &       43.15 &  12.05 &  28.71 &  27.58   &       32.76 &  12.05 \\
ATL \cite{hou2021atl}  &            52.01 &        50.94  &     59.44 &    15.64 & 36.80 &    34.38      &     42.00 &  15.64 \\
ATL$\ast$ \cite{hou2021atl}  &    56.05 &       40.83 &       57.41 & 8.52 & 37.01 &      30.21  &       43.29 & 8.52\\

\hline
SCL$-$  & 50.51  & 43.52 & 57.29 & 14.46 & 44.21  & 41.37 & 48.68 & 14.46  \\
SCL   & {\bf 59.64} & {\bf 52.70} & {\bf 67.05} & 14.90  & {\bf 47.68} & {\bf 42.05} & {\bf 52.95} & 14.90 \\
SCL (3M iters)  & {\bf 72.08} & {\bf 57.53} &  {\bf 82.47} & {\bf 18.55} & {\bf 56.19} & {\bf 46.32} & {\bf 64.50} & {\bf 18.55} \\

\hline



\end{tabular}
\end{table}

\subsection{Object Affordance Recognition}
\label{sec:affordance}

Following~\cite{hou2021atl} that has discussed average precision (AP) is more robust for evaluating object affordance, we evaluate object affordance recognition with AP on HICO-DET. Table~\ref{table:func_obj_ap} illustrates SCL largely improves SCL$-$ (without self-training) by {\bf over 9\%} on Val2017, Object365, HICO-DET under the same training iterations. SCL requires more iterations to converge, and SCL greatly improves previous methods on all datasets with 3M iterations (Please refer to Appendix D.2 for convergence analysis). Noticeably, SCL directly predicts verb rather than HOI categories, and removes the spatial branch. Thus, SCL without self-training (SCL$-$) is a bit worse than ATL. Previous approaches ignore the unknown affordance recognition. We use the released models of ~\cite{hou2021atl} to evaluate the results on novel affordance recognition. Here, affordances of novel classes (annotated by hand~\cite{hou2021atl}) are the same in the two settings. We find SCL improves the performance considerably by {\bf over 10\%} on Val2017 and HICO-DET.

\subsection{HOI Detection with Unknown Concepts}

HOI concept discovery enables zero-shot HOI detection with unknown concepts by first discovering unknown concepts and then performing HOI detection. The experimental results of HOI detection with unknown concepts are shown in Table~\ref{table:zs_unknown}. We follow~\cite{hou2020visual} to evaluate HOI detection with 120 unknown concepts in two settings: rare first selection and non-rare first selection, \ie, we select 120 unknown concepts from head and tail classes respectively. Different from \cite{hou2020visual,hou2021fcl} where the existence of unseen categories is known and the HOI samples for unseen categories are composed during optimization, HOI detection with unknown concepts does not know the existence of unseen categories. Therefore, we select top-$K$ concepts according to the confidence score during inference to evaluate the performance of HOI detection with unknown concepts (that is also zero-shot) in the default mode~\cite{chao2018learning}.

As shown in Table~\ref{table:zs_unknown}, with more selected unknown concepts according to concept confidence, the proposed approach further improves the performance on unseen categories on both rare first and non-rare first settings. Specifically, it demonstrates a large difference between rare first unknown concepts HOI detection and non-rare first unknown concepts HOI detection in Table~\ref{table:zs_unknown}. Considering that the factors (verbs and objects) of rare-first unknown concepts are rare in the training set~\cite{hou2021fcl}, the recall is very low and thus degrades the performance on unknown categories. However, with concept discovery, the results with top 120 concepts on unknown categories are improved by relatively \textbf{34.52\%} (absolutely 0.58\%) on rare first unknown concepts setting and by relatively \textbf{20.31\%} (absolutely 1.19\%) on non-rare first setting, respectively. with more concepts, the performance on unknown categories is also increasingly improved.

\begin{table*}[tp]
\small
\setlength\tabcolsep{3.5pt}
\caption{Illustration of HOI detection with unknown concepts and zero-shot HOI detection with SCL. $K$ is the number of selected unknown concepts. HOI detection results are reported by mean average precision (mAP)(\%). We also report the recall rate of the unseen categories in the top-$K$ novel concepts. ``$K$ = all" indicates the results of selecting all concepts, \ie, common zero-shot. $\ast$ means we train Qpic~\cite{tamura_cvpr2021}(ResNet-50) with the released code in zero-shot setting and use the discovered concepts of SCL to evaluate HOI detection with unknown concepts. Un indicates Unknown/Unseen, Kn indicates Known/Seen, while Rec indicates Recall.}
\label{table:zs_unknown}
\centering

\begin{tabular}{@{}lccccccccc@{}}
\hline
\multirow{2}{*}{Method} & \multirow{2}{*}{$K$} &
\multicolumn{4}{c}{Rare First}&\multicolumn{4}{c}{Non-rare First}\cr\cline{3-10}
& & Un & Kn &Full & Rec (\%)& Un & Kn &Full & Rec (\%)\cr

\hline



SCL & 0   & 1.68& 22.72 & 18.52 & 0.00 & 5.86 & 16.70 & 14.53 & 0.00\\
SCL & 120 & 2.26& 22.72 & 18.71 & 10.83 & 7.05 & 16.70 & 14.77 & 21.67\\
SCL  &  240 & 3.66 & 22.72 & 18.91 & 15.00 & 7.17 & 16.70 & 14.80 & 25.00\\
SCL & 360 & 4.09 & 22.72 & 19.00 & 15.83 & 7.91 & 16.70 & 14.94 & 30.83\\
SCL &  all & 9.64 & 22.72 & 19.78 & 100.00 & 13.30 & 16.70 & 16.02 & 100.00\\
\hline

Qpic$\ast$~\cite{tamura_cvpr2021} & 0 & 0.0 & {\bf 30.47} & 24.37 & 0.00 & 0.0 & 23.73 & 18.98 & 0.0\\
Qpic$\ast$~\cite{tamura_cvpr2021} &120 & 2.32  & 30.47 & 24.84 & 10.83 & 14.90 & 22.19 & 20.58 & 21.67 \\
Qpic$\ast$~\cite{tamura_cvpr2021} &240 & 3.35  & 30.47 & 25.04 & 15.00 & 14.90 & 22.79 & 21.22 & 25.00 \\
Qpic$\ast$~\cite{tamura_cvpr2021} & 360 & 3.72  & 30.47 & 25.12 & 15.83 & 14.91 & 23.13 & 21.48 & 30.83 \\
Qpic$\ast$~\cite{tamura_cvpr2021} & all &15.24 & 30.44 & 27.40 & 100.00 & 21.03 & 23.73 & 23.19 & 100.00\\
\hline\hline
ATL~\cite{hou2021atl} & all & 9.18 & 24.67 & 21.57& 100.00 & 18.25 & 18.78 & 18.67 & 100.00\\
FCL~\cite{hou2021fcl} & all & 13.16 & 24.23 & 22.01 & 100.00 & 18.66 & 19.55 & 19.37 & 100.00 \\
Qpic + SCL & all & {\bf 19.07} & 30.39 & {\bf 28.08} & 100.00 & {\bf 21.73} & {\bf 25.00} & {\bf 24.34} & 100.00\\

\hline
\end{tabular}
\end{table*}

We also utilize the discovered concept confidences with SCL to evaluate HOI detection with unknown concepts on Qpic~\cite{tamura_cvpr2021}. For a fair comparison, we use the same concept confidences to SCL. Without concept discovery, the performance of Qpic~\cite{tamura_cvpr2021} degrades to 0 on Unseen categories though Qpic significantly improves zero-shot HOI detection.
Lastly, we show zero-shot HOI detection (the unseen categories are known) in Table~\ref{table:zs_unknown} (Those rows where $K$ is all). We find that SCL significantly improves Qpic, and \textit{forms a new state-of-the-art} on zero-shot setting though we merely use ResNet-50 as backbone in Qpic. We consider SCL improves the detection of rare classes (include unseen categories in rare first and seen categories in non-rare first) via stating the distribution of verb and object. See Appendix F for more analysis, \eg, SCL improves Qpic particularly for rare categories on Full HICO-DET.

\subsection{Visualization}

\begin{figure}[!ht]
\centering
    \includegraphics[width=0.75\linewidth]{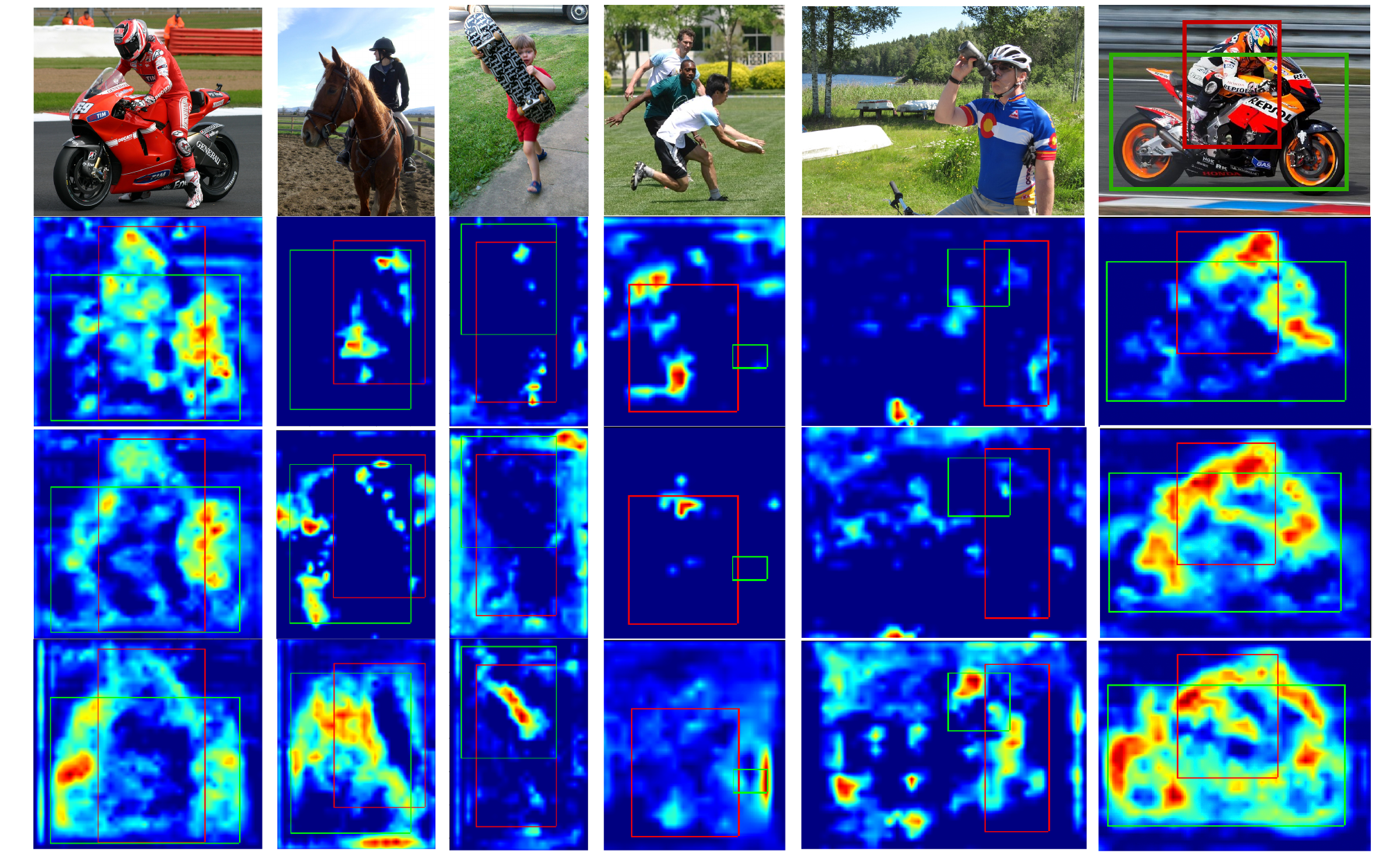}
    \caption{A visual comparison of recent methods using the Grad-CAM~\cite{selvaraju2017gradcam} tool. The first row is input image, the second row is baseline without compositional approach, the third row is VCL~\cite{hou2020visual} and the last row is the proposed SCL. We do not compare with ATL~\cite{hou2021atl}, since that ATL uses extra training datasets. Here, we compare all models using the same dataset. }
    \label{fig:vis_grad_cam}
    \vspace{-8mm}
\end{figure}

 Figure~\ref{fig:vis_grad_cam} illustrates the Grad-CAM under different methods. We find the proposed SCL focus on the details of objects and small objects, while the baseline and VCL mainly highlight the region of human and the interaction region, \eg, SCL highlights the details of the motorbike, particularly the front-wheel (last row). Besides, SCL also helps the model via emphasizing the learning of small objects (\eg, frisbee and bottle in the last two columns), while previous works ignore the small objects. This demonstrates SCL facilitates affordance recognition and HOI concept discovery via exploring more details of objects. A similar trend can be observed in Appedix G (Qpic+SCL).




\section{Conclusion}
\label{sec:con}

We propose a novel task, Human-Object Interaction Concept Discovery, which aims to discover all reasonable combinations (\ie, HOI concepts) between verbs and objects according to a few training samples  of known HOI concepts/categories. Furthermore, we introduce a self-compositional learning or SCL framework for HOI concept discovery. SCL maintains an online updated concept confidence matrix, and assigns pseudo labels according to the matrix for all composite HOI features, and thus optimize both known and unknown composite HOI features via self-training. SCL facilitates affordance recognition of HOI model and HOI concept discovery via enabling the learning on both known and unknown HOI concepts. Extensive experiments demonstrate SCL improves HOI concept discovery on HICO-DET and V-COCO and object affordance recognition with HOI model, enables HOI detection with unknown concepts, and improves zero-shot HOI detection. \\
{\bf Acknowledgments} Mr. Zhi Hou and Dr. Baosheng Yu are supported by ARC FL-170100117, DP-180103424, IC-190100031, and LE-200100049.

\clearpage
%
%
\bibliographystyle{splncs04}
\bibliography{egbib}

\appendix

\section{Detailed Analysis for the Motivation}
\label{sec:detailed}

Actually, after we generate the composite HOI features, we have features for both known and unknown concepts. We merely know the HOI features of the known concepts are existing, while we do not know whether the HOI features of unknown concepts are reasonable or not. This actually fall into a typical semi-supervised learning, in which part of samples are labeled (known). Therefore, inspired by the popular semi-supervised learning method, we propose to design a self-training strategy with pseudo labels.

SCL largely improves concept discovery. At first, during training, SCL involves both HOI instances from known or unknown concepts (via pseudo-labeling). Another important thing is that SCL uses both positive and negative unknown concepts, which prevents the model from only fitting the verb patterns. {\small For example, the classifier may predict a reasonable concept for the verb ``eat" regardless of the object representation, if there are no negative unknown concepts, \eg, ``eat TV". Lastly, as shown in Figure~\ref{fig:convergence}, SCL also reduces the risk of overfitting known concepts compared with ATL. \eg, we observe high confidence for the novel concept "squeeze banana"(sort in 2027) in SCL, while the confidence of "squeeze banana" is merely 0.0017 (sort in 7554) in ATL.}

\section{Annotation}
\label{sec:annotation}

In order to evaluate the proposed method, we manually annotate the novel concepts for both HICO and V-COCO dataset. Specifically, we annotate the concepts that people can infer from existing concepts. The final set of concepts are provided in the supplemental material. 

Statistically, there are about 1.3\% and 1.9\% mislabeled pairs on HICO-DET and V-COCO, respectively. Meanwhile, there are about 1.7\% and 1.1\% unlabeled pairs (including ambiguous verbs) on the remaining categories of HICO-DET and V-COCO.

To evaluate the effect of annotation quality of concept annotation on HOI concept discovery, we illustrate the result of different models with different annotations. We compare two versions of annotations, both of which are provided in supplemental materials. Specifically, the file ``label\_hoi\_concept.csv" is the worse version, while ``label\_hoi\_concept\_new.csv" is the refined version. Table~\ref{table:discover_anno} shows SCL even achieves better performance when evaluate SCL with better annotation, while the performance of baseline is not improved. This experiments together with Table 1 in the main paper show the quality of current annotation is enough for the evaluation of the proposed method.

\begin{table}[!tp]
\small
\caption{The performance of the proposed method for HOI concept discovery under different annotations. Better Annotation indicates we remove some wrongly labeled concepts in annotation. We report all performance using the average precision (AP) (\%). UC means unknown concepts and KC means known concepts. SCL means self-compositional learning. SCL$-$ means online concept discovery without self-training. 
}
\label{table:discover_anno}
\centering
\small
\begin{tabular}{@{}lccc@{}}
\hline
Method & Better Annotation & UC & KC \cr
\hline
SCL$-$ & & 22.36 & 83.04  \\
SCL & &   33.26  & 93.06   \\
\hline
SCL$-$ & \checkmark & 22.25 & 83.04  \\
SCL & \checkmark & {\bf 33.58} & {\bf 92.65}  \\


\hline
\end{tabular}
\end{table}

\section{Qualitative illustration}
\label{sec:quan}

We also illustrate the discover concepts in this Section. Here, we choose the concepts after removing the known concepts from the prediction list because the confidence of known concepts in the prediction of SCL is usually very higher. We choose 5 concepts with high confidence and 5 concepts with low confidence to illustrate. Table~\ref{table:concepts_vis} shows the discovered concepts in SCL are usually more reasonable. We provide the full prediction list with confidence in ``result\_conf\_SCL.txt" in supplementary materials.
\begin{table}[!tp]
\small
\caption{The illustration of discovered concepts.}
\label{table:concepts_vis}
\centering
\small

\begin{tabular}{@{}l|p{0.4\textwidth}|p{0.4\textwidth}@{}}
\hline 
Method & Concepts with high confidence & Concepts with low confidence \cr
\hline
SCL$-$ & type\_on sink,inspect refrigerator,feed suitcase, inspect chair,carry stop\_sign & zip zebra, sign dog, chase broccoli, set parking\_meter, tag teddy\_bear\\
\hline
SCL & ride bear, board truck, carry bowl, wash fire\_hydrant, hop\_on motorcycle & zip zebra, flush parking\_meter,
stop\_at hair\_drier, stop\_at microwave \\

\hline
\end{tabular}
\end{table}

\section{Ablation Studies}
\label{sec:ab}

\subsection{Modules}

We conduct ablation studies on three modules: verb auxiliary loss~\cite{hou2021fcl}, union verb~\cite{hou2020visual}, and spatial branch~\cite{gao2018ican}. Union verb indicates that we extract verb representation from the union box of human and object. When we remove the union verb representation, we directly extract verb representation from the human bounding box; In our experiment, we remove the spatial branch. Here, we demonstrate we achieve better performance without the spatial branch.

{\bf Spatial branch.} We remove the spatial branch in~\cite{gao2018ican}, which is very effective for HOI detection. We find that the spatial branch degrades the performance of HOI concept discovery: the performance of HOI concept discovery increases from 32.56\% to 33.26\% without spatial branch, as shown in Table~\ref{table:ab_modules}. We thus remove spatial branch.

{\bf Verb auxiliary loss.} We follow~\cite{hou2021fcl} to utilize a verb auxiliary loss to regularize verb representations. As shown in Table~\ref{table:ab_modules}, the model without using a verb auxiliary loss drops by nearly 3\% on unseen concepts, which demonstrates the importance of verb auxiliary loss for HOI concept discovery.

{\bf Union verb.} Table~\ref{table:ab_modules} demonstrates that extracting verb representation from union box is of great importance for HOI concept discovery. When we extract verb representation from human bounding box, the result of HOI concept discovery apparently drops from 32.56\% to 28.30\%. 

Though verb auxiliary loss and union verb representation are very helpful for concept discovery, the performance without the two strategies still outperform our baseline, \ie, online concept discovery without self-training.

\begin{table}[tp]
\caption{Ablation studies of different modules on HICO-DET. UC means unknown concepts and KC means known concepts. Verb aux loss means Verb auxiliary loss (\ie, binary cross entropy loss). Results are reported by average precision (\%).}
\label{table:ab_modules}
\setlength\tabcolsep{4pt}
\centering

\begin{tabular}{@{}ccccc@{}}
\hline 
Spatial branch & Verb aux loss & Union Verb  & UC & KC\cr

\hline
\checkmark & \checkmark & \checkmark & 32.56 & 94.39 \\
- & \checkmark & \checkmark & {\bf 33.26} & 93.06 \\
\checkmark & - & \checkmark &  29.56 & 93.36 \\
\checkmark & \checkmark & - & 28.30 & 94.27 \\


\hline 
\end{tabular}

\end{table}

\subsection{Convergence Analysis}
\begin{figure}
    \centering
    \includegraphics[width=\linewidth]{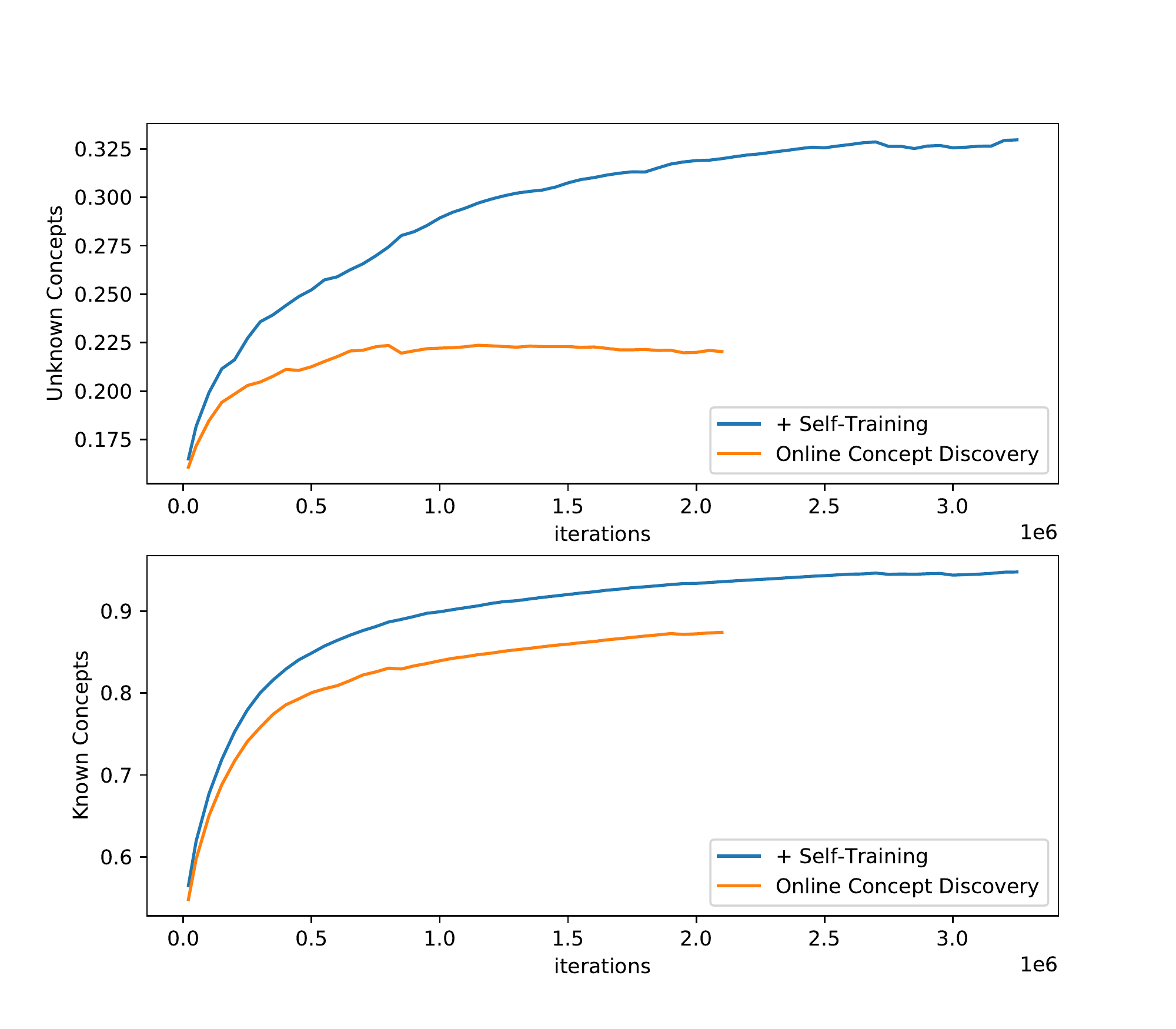}
    \caption{Illustration of the convergence with self-training strategy.}
    \label{fig:convergence}
\end{figure}


To some extent, the self-training approach makes use of all composite HOIs, and thus significantly enriches the training data. As a result, the self-training strategy usually requires more iterations to converge to a better result. Figure~\ref{fig:convergence} illustrates the comparison of convergence between online concept discovery and self-training. For online concept discovery, we observe that the model begins to overfit the known concepts after 2,000,000 iterations, and we thus have an early stop during the optimization. We notice that the result on unknown concepts of self-training increases to 32.\%, while the baseline (\ie, online concept discovery) begins to overfit after 800,000 iterations. This might be because the self-training utilizes all composite HOIs including many impossible combinations (\ie, negative samples for HOI concept discovery).

\begin{table}[tp]
\setlength\tabcolsep{4pt}
\small
\caption{Ablation studies of hyper-parameters on V-COCO. UC means unknown concepts and KC means known concepts. Results are reported by average precision (\%).}
\label{table:ab_hyper}
\centering

\begin{tabular}{@{}l|c|cccccc@{}}
\hline 
$\lambda_3$ & 0.5 & 0.5 & 0.5 & 0.25 & 1. & 2. & 4.\\
\hline
$T$ & 1 & 2 & 0.5 & 1. & 1. & 1. & 1.\\
\hline
UC (\%) & 29.52 & 28.60 & 29.69 & 28.06 & 29.94 & {\bf 31.33} & 29.78\\
KC (\%) & 97.57 & 96.76 & 97.57 & 95.32 & 97.87 & 97.81 & 97.94\\

\hline 
\end{tabular}

\end{table}

\subsection{Hyper-parameters}

In the main paper, we have several hyper-parameters (\ie $\lambda_1$, $\lambda_2$, $\lambda_3$, $T$, where $\lambda_1=2.$, $\lambda_2=0.5$, $\lambda_3=0.5$ and $T=1.$). For $\lambda_1$ and $\lambda_2$, we follow the settings in ~\cite{hou2020visual}. For $\lambda_3$ and $T$, we perform ablation studies on V-COCO as shown in Table~\ref{table:ab_hyper}. We notice that both $T$ and $\lambda_3$ have an important effect on the HOI concept discovery. As shown in Table~\ref{table:ab_hyper}, the performance increases from 29.52\% to {\bf 31.33\%} on unseen concepts when we set $\lambda_3 = 2.$, which is much better than the results reported in the main paper. This also illustrates that $\mathcal{L}_d$ is more important than $\mathcal{L}_{CL}$ for HOI concept discovery.

In our experiment, we apply the temperature $T$ to predictions. As shown in Table~\ref{table:ab_hyper}, we find that when $T$ decreases to 0.5, the performance also slightly increases from 29.52\% to 29.69\%. Thus, we further conduct ablation experiments on $T$ in Table~\ref{table:ab_hyper_t}. Specifically, to quickly evaluate the effect of $T$, we remove spatial branch and run all experiments with 1,000,000 iterations. Noticeably, when we set $T=0.25$, the performance on concept discovery further increases from 30.36\% to {\bf 33.66\%}, which indicates a smaller temperature helps HOI concept discovery. In our experiments, we also find this result further increases to over 35.\% when $T=0.5$ after convergence, which is much better than the result (33.26\%) of $T=1$. This might be because smaller temperature is less sensitive to noise data, since composite HOIs can be regard as noise data.

\begin{table}[tp]
\caption{Ablation studies of hyper-parameter $T$ on HICO-DET. Here, we run all experiments with only 1,000,000 iterations and remove the spatial branch to evaluate $T$. UC means unknown concepts and KC means known concepts. Results are reported by average precision (\%).}
\label{table:ab_hyper_t}
\centering

\begin{tabular}{@{}lc|c|ccc@{}}
\hline 

$T$ & 2 & 1 & 0.5 & 0.25 & 0.125\\
\hline 
UC (\%)  & 27.15 & 30.36 & 33.54 & {\bf 33.66} &  33.25 \\
KC (\%)  & 85.53 & 88.72 & 91.71 & 93.62 & {\bf 94.32}\\
\hline 
\end{tabular}

\end{table}

\begin{table}[tp]
\small
\caption{Illustration of normalized pseudo labels on HICO-DET and V-COCO. Experiments results are reported by average precision (\%). Here, the SCL model uses spatial branch.}
\label{table:ab_normaliz}
\centering
\small
\begin{tabular}{@{}lcccc@{}}
\hline 
 \multirow{2}{*}{Method} &
\multicolumn{2}{c}{HICO-DET}&\multicolumn{2}{c}{V-COCO}\cr\cline{2-5}
& UC (\%) & KC (\%) &  UC (\%) & KC (\%) \cr
\hline 

SCL & {\bf 32.56} & {\bf 94.39} & {\bf 29.52} & {\bf 97.57}\\
w/o normalization & 32.30 &  94.2 & 29.32 & 97.93 \\

\hline 
\end{tabular}
\end{table}

\subsection{Normalization for Pseudo-labels}
In our experiment, we normalize the confidence matrix for pseudo-labels. Table~\ref{table:ab_normaliz} illustrates the normalization approach has a slight effect on the concept discovery performance.

\section{HOI Detection with Unknown Concepts}
\label{sec:novel_obj}

\subsection{Additional Comparisons}
Table~\ref{table:zs_unknown_app} demonstrates SCL consistent improves the baseline (\ie, SCL without Self-Training). Here, we use the same concepts for a fair comparison. Thus, the recall is the same. Meanwhile, Table~\ref{table:zs_unknown_app} also shows Self-Training effectively improves the HOI detection. when we select all concepts to evaluate HOI detection, it is common zero-shot HOI detection, \ie, all unseen classes are known. Particularly, for application, one can directly detect unknown concepts with concept discovery from the model itself, \eg, Qpic~\cite{tamura_cvpr2021}. Here, we mainly demonstrate different methods with the same concept confidence for a fair comparison.

\begin{table*}[tp]
\small
\setlength\tabcolsep{3pt}
\caption{Illustration of HOI detection with unknown concepts and zero-shot HOI detection with SCL. $K$ is the number of selected unknown concepts. HOI detection results are reported by mean average precision (mAP)(\%). We also report the recall of the unseen categories in the top-$K$ novel concepts. $K$ = all indicates the results of selecting all concepts, \ie, common zero-shot. $\ast$ means we train Qpic~\cite{tamura_cvpr2021}(ResNet-50) with the released code in zero-shot setting and use the discovered concepts of SCL to evaluate HOI detection with unknown concepts.}
\label{table:zs_unknown_app}
\centering

\begin{tabular}{@{}lccccccccc@{}}
\hline 
\multirow{2}{*}{Method} & \multirow{2}{*}{$K$} &
\multicolumn{4}{c}{Rare First}&\multicolumn{4}{c}{Non-rare First}\cr\cline{3-10}
& &Unknown &Known&Full & Recall (\%)&Unknown &Known&Full & Recall (\%)\cr

\hline



Baseline & 0 & 1.68& 22.10 & 18.52 & 0.00 & 5.86 & 16.30 & 14.21 & 0.00\\
Baseline & 120 & 3.06& 22.10 & 18.29 & 10.83 & 6.16 & 16.30 & 14.27 & 21.67\\
Baseline  &  240 & 3.28 & 22.10 & 18.34 & 13.33 & 6.90 & 16.30 & 14.42 & 25.00\\
Baseline & 360 & 3.86 & 22.10 & 18.45 & 15.83 & 7.29 & 16.30 & 14.50 & 30.83\\
Baseline & all & 9.62 & 22.10 & 19.61 & 100.00 & 12.82 & 16.30 & 15.60 & 100.00\\
\hline
SCL & 0   & 1.68& 22.72 & 18.52 & 0.00 & 5.86 & 16.70 & 14.53 & 0.00\\
SCL & 120 & 2.26& 22.72 & 18.71 & 10.83 & 7.05 & 16.70 & 14.77 & 21.67\\
SCL  &  240 & 3.66 & 22.72 & 18.91 & 15.00 & 7.17 & 16.70 & 14.80 & 25.00\\
SCL & 360 & 4.09 & 22.72 & 19.00 & 15.83 & 7.91 & 16.70 & 14.94 & 30.83\\

SCL &  all & 9.64 & 22.72 & 19.78 & 100.00 & 13.30 & 16.70 & 16.02 & 100.00\\


\hline 
\end{tabular}
\end{table*}

\subsection{Novel Objects}

In the main paper, we illustrate the result on two compositional zero-shot settings. Here, we further illustrate the effectiveness of HOI concept discovery for novel object HOI detection. Novel object HOI detection requires to detect HOI with novel objects, \ie, the object of an unseen HOI is never seen in the HOI training set. We follow~\cite{hou2021atl} to select 100 categories as unknown concepts. The remaining categories do not include the objects of unseen categories. Here we use a unique object detector to detect objects. To enable the novel object HOI detection and novel object HOI concept discovery, we follow~\cite{hou2021atl} to incorporate external objects (\eg COCO~\cite{lin2014microsoft}) to compose novel object HOI samples. Specifically, we only choose the novel types of objects from COCO~\cite{lin2014microsoft} as objects images in the framework~\cite{hou2021atl} for novel object HOI detection with unknown concepts.

Table~\ref{table:zs_unknown_object} demonstrates concept discovery largely improves the performance on unseen category from 3.92\% to {\bf 11.41\%} (relatively by 191\%) with top 100 unknown concepts. We meanwhile find the recall increases to 41.00\% with only the top 100 unknown concepts. Nevertheless, when we select all unknown concepts, the performance on unseen category is 17.19\%. This shows we should improve the performance of concept discovery.

\begin{table}[tp]
\small
\caption{Illustration of the effectiveness of HOI concept discovery for HOI detection with unknown concepts (novel objects). $K$ is the number of selected unknown concepts. HOI detection results are reported by mean average precision (mAP)(\%). Recall is evaluated for the unseen categories under the top-$k$ novel concepts. The last row indicates the results of selecting all concepts.}
\label{table:zs_unknown_object}
\centering

\begin{tabular}{@{}lccccc@{}}
\hline 
{$K$}&Unseen&Seen&Full & Recall (\%)\cr

\hline 
0   & 3.92 & 19.45 & 16.86 & 0.00  \\
\hline
100 & 11.41 & 19.45 & 18.11 & 41.00  \\
200 & 12.40 &  19.45 & 18.28 & 48.00  \\
300 & 13.52 &  19.45 & 18.46 & 52.00  \\
400 & 13.52 &  19.45 & 18.46 & 52.00  \\
500 & 13.91 &  19.45 & 18.53 & 56.00  \\
600 & 13.91 &  19.45 & 18.53 & 56.00  \\
\hline
all & 17.19 &  19.45 & 19.07 & 100.00 \\
\hline 
\end{tabular}
\end{table}

\begin{table}[!tp]
\small
\caption{Additional Comparison on HOI concept discovery. We report all performance using the average precision (AP) (\%). UC means unknown concepts and KC means known concepts. SCL means self-compositional learning. SCL$-$ means online concept discovery without self-training. 
SCL (COCO) means we train the network via composing between verbs from HICO and objects from COCO 2014 training set.
}
\label{table:discover_app}
\centering
\small
\begin{tabular}{@{}lcccc@{}}
\hline
 \multirow{2}{*}{Method} &
\multicolumn{2}{c}{HICO-DET}&\multicolumn{2}{c}{V-COCO}\cr\cline{2-5}
& UC (\%) & KC (\%) &  UC (\%) & KC (\%) \cr
\hline
Random & 12.52 & 6.56 & 12.53 & 13.54\\
language embedding & 16.08 & 29.64 & - & - \\

Re-Training &  26.09 & 50.32 & - & -\\
\hline
SCL$-$ (COCO) & 17.01 & 55.50 &  26.04 & 81.47 \\
SCL (COCO) & 31.92 & 86.43 & 27.90 & 90.04 \\
\hline
SCL$-$ & 22.36 & 83.04 & 26.64 & 95.59\\


SCL & {\bf 33.26} & {\bf 93.06} & {\bf 29.52} & {\bf 97.57}\\


\hline
\end{tabular}
\end{table}

\begin{table}[!ht]
\caption{Illustration of the effectiveness of self-training on HOI detection based on ground truth box. Results are reported by mean average precision (\%). }
\label{table:hoi}
\centering

\begin{tabular}{@{}ccccc@{}}
\hline
Method & Full & Rare & NonRare\cr

\hline
SCL & {\bf 42.92} & {\bf 36.60} & {\bf 44.81} \\
w/o Self-Training & 42.66 & 35.81 & 44.70 \\

\hline
\end{tabular}

\end{table}

\begin{table}[!ht]
\caption{Illustration of the effectiveness of self-training for Qpic (ResNet-50). Results are reported by mean average precision (\%). $*$ means we use the released code to reproduce the results for a fair comparison. S1 means Scenario 1, while S2 means Scenario 2.}
\label{table:hoi_qpic}
\centering
\setlength\tabcolsep{3.5pt}
\begin{tabular}{@{}cccccc@{}}
\hline
 \multirow{2}{*}{Method} &
\multicolumn{3}{c}{HICO-DET}&\multicolumn{2}{c}{V-COCO}\cr\cline{2-6}

 & Full & Rare & NonRare & S1 & S2\cr

\hline
GGNet~\cite{zhong2021glance} & 23.47 & 16.48 & 25.60 & - & 54.7 \\
ATL~\cite{hou2021atl} & 23.81 & 17.43 & 25.72 & - & - \\
HOTR~\cite{kim2021hotr} & 25.10 & 17.34 & 27.42 & 55.2 & {\bf 64.4}\\
AS-Net\cite{chen2021reformulating} & 28.87 & 24.25 & 30.25 & - & 53.9\\
Qpic~\cite{tamura_cvpr2021} & 29.07 & 21.85 & 31.23 & 58.8 & 61.0\\
\hline
Qpic*~\cite{tamura_cvpr2021} & 29.19 & 23.01 & 31.04 & 61.29 & 62.10\\
Qpic + SCL & {\bf 29.75} & {\bf 24.78} & {\bf 31.23} & {\bf 61.55} & {\bf 62.38} \\
\hline
\end{tabular}

\end{table}

\section{HOI Detection}
\label{sec:hoi_det}

{\bf One-Stage Method.} We also evaluate SCL on Qpic~\cite{tamura_cvpr2021}, \ie, the state-of-the-art HOI detection method based on Transformer, for HOI detection. Code is provided in~\url{https://github.com/zhihou7/SCL}. We first obtain concept confidence similar as Section 3.3.2 in the main paper. Denote $\mathbf{\hat{Y}}_v \in R^{N \times N_v}$ as verb predictions,  $\mathbf{\hat{Y}}_o \in R^{N \times N_o}$ as verb predictions, we obtain concept predictions $\mathbf{\hat{Y}}_{h}$ as follows,
\begin{equation}
\label{eq:qpic_pred}
\mathbf{\hat{Y}}_{h} = \mathbf{\hat{Y}}_v \otimes \mathbf{\hat{Y}}_o.
\end{equation}

Then, we update $M$ according to Equation 2 and Equation 3 in the main paper. After training, we evaluate HOI concept discovery with $M$.

For self-training on Qpic~\cite{tamura_cvpr2021}, we use $M$ to update the verb label $\mathbf{Y}_v\in R^{N\times N_v}$ for annotated HOIs. Here, we do not have composite HOIs because Qpic has entangled verb and object predictions, and we update verb labels with $\mathbf{M}$. Specifically, given an HOI with a verb labeled as $y_v\in R^N_v$ and an object labeled as $y_o \in R^N_o$, where $0\leq y_o < N_o$ denotes the index of object category, we update $y_v$ as follows,

\begin{equation}
\label{eq:qpic_pred1}
\widetilde{y}_v = max(y_v+\mathbf{M}(:, y_o), 1)
\end{equation}

where $max$ means we clip the value to 1 if the value is larger than 1. Then, we obtain pseudo verb label $\widetilde{y}_v$ to optimize the samples of the HOI similar as Equation 7 (here, we only have annotated HOI samples). We think the running concept confidence $\mathbf{M}$ have {\bf implicitly counted the distribution of verb and object in the dataset}. Meanwhile, the denominator in Equation 2 can also normalize the confidence according to the frequency, and thus ease the long-tailed issue. Thus, with the pseudo labels constructed from $\mathbf{M}$, we can re-balance the distribution of the dataset, which is a bit similar to re-weighting strategy~\cite{byrd2019effect,cui2019class}. However, SCL does not require to set the weights for each class manually.

Table~\ref{table:hoi_zs} demonstrates SCL greatly improves Qpic on Unseen category on rare first zero-shot detection, while SCL significantly facilitates rare category on non-rare first zero-shot detection. In Full HOI detection on HICO-DET, Table~\ref{table:hoi_qpic} shows SCL largely facilitates HOI detection on rare category. Particularly, the seen category in rare first setting includes 120 rare classes, while the seen category in non-rare first setting only includes 18 classes (all rare classes are in unseen category in non-rare first setting). Thus, SCL actually improves HOI detection for rare category. We think the concept confidence matrix internally learns the distribution of verb and objects and in the dataset. \eg, given an object, $\mathbf{M}$ illustrates the corresponding verb distribution.

\begin{table}[!ht]
\caption{Zero-Shot HOI detection based on Qpic. Results are reported by mean average precision (\%). Here, we split the classes of HOI into four categories in zero-shot setting, \ie, Seen are categorized into rare and non-rare. }
\label{table:hoi_zs}
\setlength\tabcolsep{3.5pt}
\centering

\begin{tabular}{@{}cccccc@{}}
\hline
Method & Unseen & Rare & NonRare & Full\cr

\hline
Qpic~\cite{tamura_cvpr2021} (non-rare first)& 21.03 & 19.12& 25.59 & 23.19 \\
Qpic+SCL (non-rare first) & {\bf 21.73} & {\bf 22.43} & {\bf 26.03} & {\bf 24.34} \\
\hline
Qpic~\cite{tamura_cvpr2021} (rare first)& 15.24 & 16.72 & 30.98 & 27.40 \\
Qpic+SCL (rare first) & {\bf 19.07} & 16.19 & 30.89 & {\bf 28.08} \\

\hline
\end{tabular}

\end{table}

{\bf Two-Stage method.} Considering the HOI concept discovery is mainly based on two-stage HOI detection approaches~\cite{hou2020visual}, it is direct and simple to evaluate the performance of self-training on HOI detection. Table~\ref{table:hoi} demonstrates the HOI detection results on ground truth boxes. Noticeably, we directly predict the verb category, rather than HOI category. Thus, the baseline of HOI detection (\ie visual compositional learning~\cite{hou2020visual}) is a bit worse. We can find self-training also slightly improves the performance, especially on rare category.

\section{Visualization}
\label{sec:visualize}

In this section, we provide more visualized illustrations.

\begin{figure*}
\centering
    \includegraphics[width=0.7\textwidth]{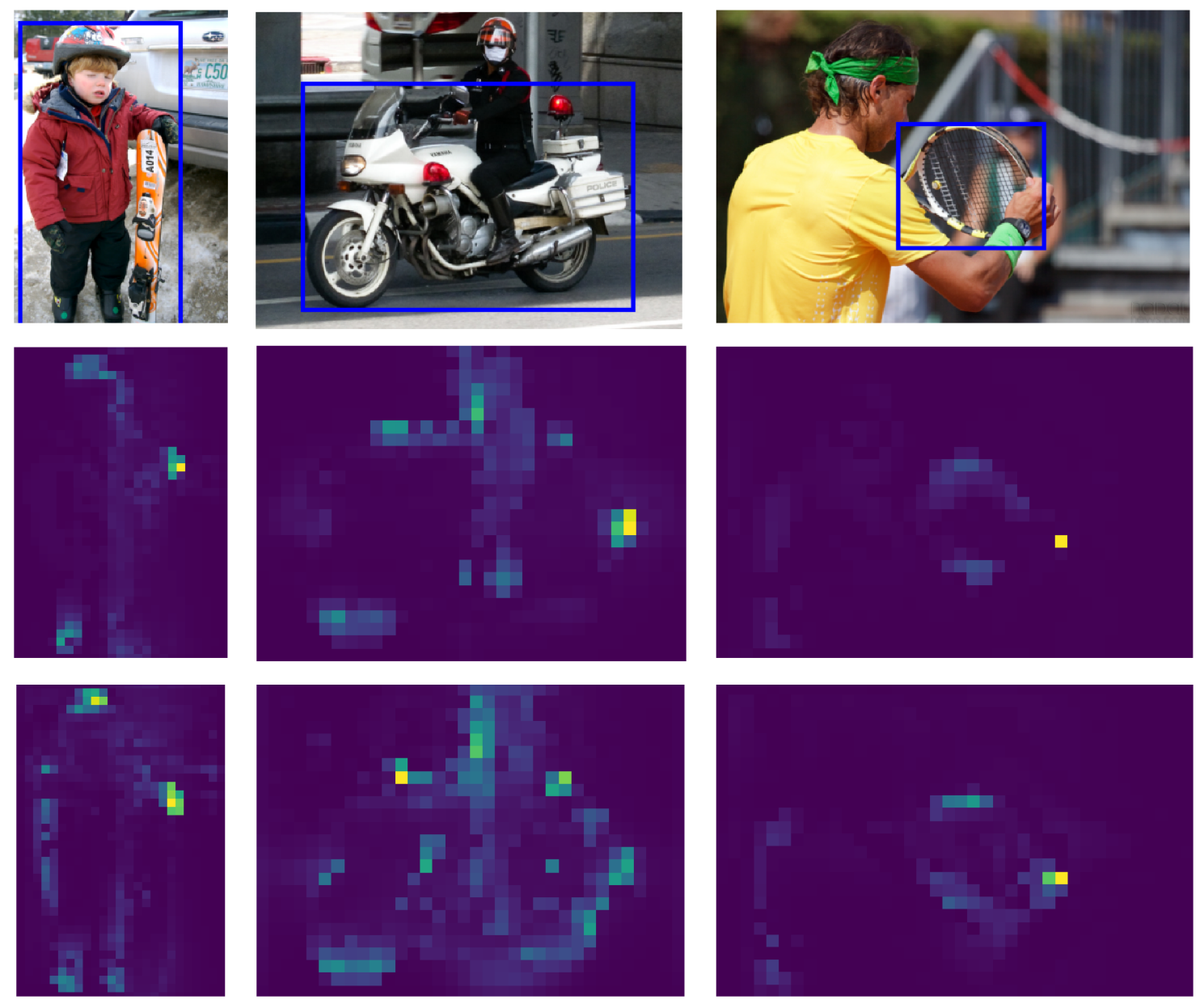}
    \caption{Visualized Illustration of SCL+Qpic and Qpic~\cite{tamura_cvpr2021}.}
    \label{fig:vis_comp_qpic}
\end{figure*}

{\bf More Grad-CAM Visualizations} 
Figure~\ref{fig:vis_comp_qpic} demonstrates the visualization of Qpic and Qpic+SCL: the second row is Qpic and the third row is Qpic+SCL, where we observe a similar trend to the Gram-CAM illustration in main paper.

\begin{figure*}
    \includegraphics[width=.9\textwidth]{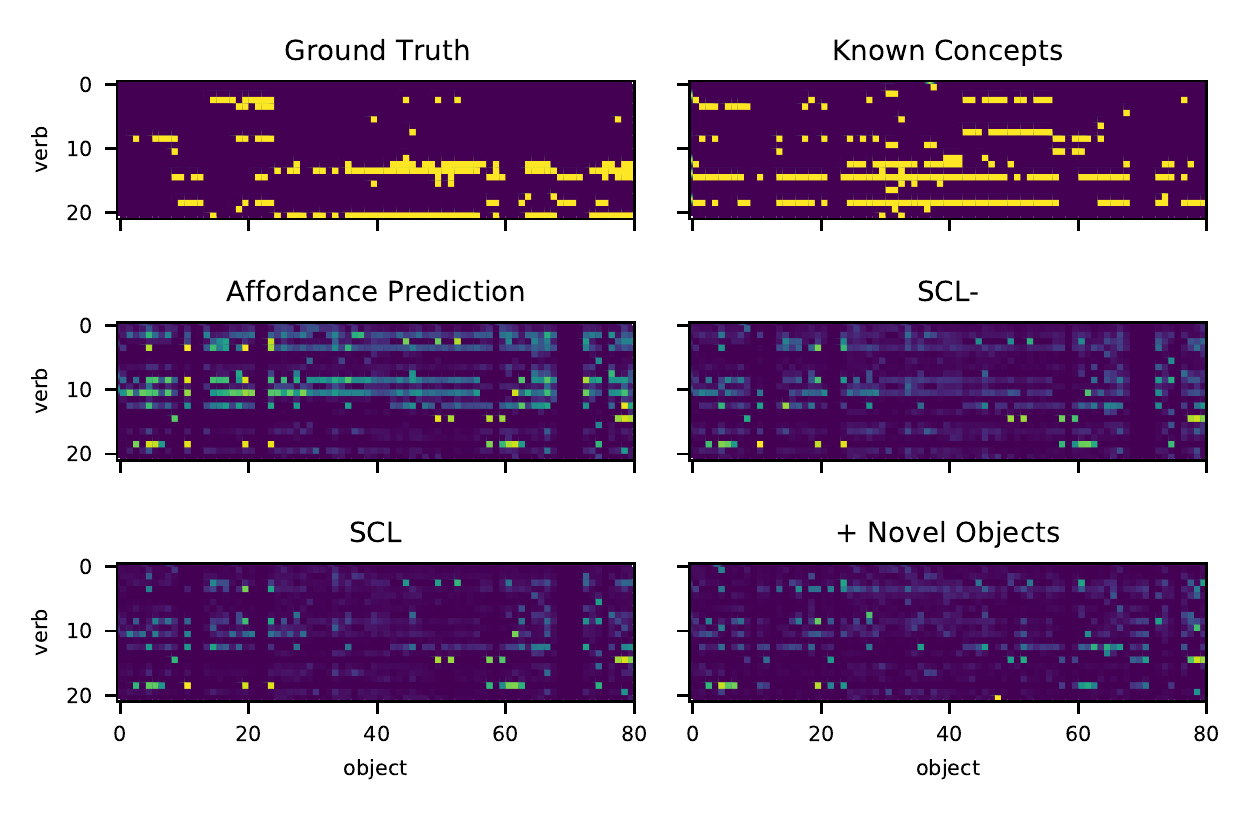}
    \caption{Visualized Comparison of different methods on V-COCO dataset. The column is the object classes and the row represents the verb classes. Known Concepts are the concepts that we have known. SCL$-$ means online concept discovery without self-training. For better illustration, we filter out known concepts in proposed methods. ``+ Novel Objects" means self-training with novel object images.}
    \label{fig:vis_comp}
\end{figure*}

{\bf Concept Visualization.} We illustrate the visualized comparisons of concept discovery in Figure~\ref{fig:vis_comp}. According to the ground truth and known concepts, we find some verb (affordance) classes can be applied to most of objects (the row is highlighted in the ground truth figure). This observation is reasonable because some kinds of actions can be applied to most of objects in visual world, \eg, hold. As shown in Figure~\ref{fig:vis_comp}, there are many false positive predictions in the results of affordance prediction, and affordance prediction tends to overfit the known concepts, especially those with frequently appeared verbs. Methods of online HOI concept discovery on V-COCO have fewer false positive predictions compared to affordance prediction. However, the two methods tend to predict concepts composed of frequent verbs in known concepts due to the verb and object imbalance issues in HOI dataset \cite{hou2021fcl}. Particularly, the false positive predictions are largely eased with self-training (\eg, the top right region). In addition, the blank columns in Figure~\ref{fig:vis_comp} are because there are only 69 objects in V-COCO training set, and we can ease it via training network with additional object images~\cite{hou2021atl} as illustrated in the last figure of Figure~\ref{fig:vis_comp}. See more visualized results on HICO-DET and V-COCO in the supplemental material. Particularly, we further notice there are dependencies between verb classes (See verb dependency analysis).

\section{Additional Concept Discovery Approaches}
\label{sec:cd_exp}

We provide More comparisons in this Section. For a fair comparison with ATL~\cite{hou2021atl} (\ie, affordance prediction), we use the same number of verbs (21 verbs) on V-COCO. The code includes how to convert V-COCO to 21 verbs, \ie merge ``\_instr'' and ``\_obj'' and remove actions without object (\eg, stand, smile, run). 

{\bf Language embedding baseline.} In the main paper, we illustrate a random baseline. Here we further illustrate the results with language embedding~\cite{pennington2014glove}. Different from extracting verb/object features from real HOI images, we use the corresponding language embedding representations of verb/object as input, \ie discovering concepts from language embedding. Table~\ref{table:discover_app} shows the performance is just a bit better than random result, and is much worse than online concept discovery. Similar to the main paper, when we evaluate the unknown concepts, we mask out the known concepts to avoid the disturbance from known concepts.

{\bf Re-Training.} We first train the HOI model via visual compositional learning~\cite{hou2020visual}, and then predict the concept confidence. Next, we use the predicted concept confidence to provide pseudo labels for the composite HOIs. Table~\ref{table:discover_app} shows the performance of Re-Training is worse than SCL. 

{\bf With COCO dataset.} Table~\ref{table:discover_app} also demonstrates the baseline (SCL$-$) with COCO datasets has poor performance on concept discovery. We think it is because the domain shift between COCO dataset and HICO-DET dataset. However, SCL still achieves significant improvement on concept discovery.

{\bf Qpic+SCL.} The details are provided in Section D.

\end{document}